\documentclass[10pt]{article}
\usepackage{amssymb,amsfonts,amstext,amsmath,amsthm}
\usepackage{graphicx}
\usepackage[round]{natbib}
\usepackage{authblk}

\newcommand{\be}{{\mathbf e}}
\newcommand{\bff}{{\mathbf f}}

\newcommand{\bx}{{\mathbf x}}
\newcommand{\btheta}{{\boldsymbol\theta}}

\newcommand{\calL}{{\mathcal L}}
\newcommand{\calR}{{\mathcal R}}
\newcommand{\calX}{{\mathcal X}}
\newcommand{\calY}{{\mathcal Y}}
\newcommand{\bbE}{{\mathbb E}}
\newcommand{\bbR}{{\mathbb R}}

\title{A Surrogate Loss Function for Optimization of $F_\beta$ Score in Binary Classification with Imbalanced Data}

\author[1,2]{Namgil Lee\thanks{namgil.lee@kangwon.ac.kr}}
\author[2,3]{Heejung Yang\thanks{heejyang@kangwon.ac.kr}}
\author[2]{Hojin Yoo\thanks{hojinyoo@bionsight.com}}
\affil[1]{Department of Information Statistics, 
	Kangwon National University, Chuncheon, Gangwon 24341, 
	Republic of Korea}
\affil[2]{Bionsight, Inc., Gangwondaehak-gil 1, Chuncheon, Gangwon 24341, Republic of Korea}
\affil[3]{Department of Pharmacy, 
	Kangwon National University, Gangwondaehak-gil 1, 
	Chuncheon, Gangwon 24341, Republic of Korea}

\date{}

\begin{document}

\maketitle

\begin{abstract}
	The $F_\beta$ score is a commonly used measure of classification performance, which plays crucial roles in classification tasks with imbalanced data sets. However, the $F_\beta$ score cannot be used as a loss function by gradient-based learning algorithms for optimizing neural network parameters due to its non-differentiability. 
	On the other hand, commonly used loss functions such as the binary cross-entropy (BCE) loss are not directly related to performance measures such as the $F_\beta$ score, so that neural networks optimized by using the loss functions may not yield optimal performance measures. 	
	In this study, we investigate a relationship between classification performance measures and loss functions in terms of the gradients with respect to the model parameters. Then, we propose a differentiable surrogate loss function for the optimization of the $F_\beta$ score. 
	We show that the gradient paths of the proposed surrogate $F_\beta$ loss function approximate the gradient paths of the large sample limit of the $F_\beta$ score. 
	Through numerical experiments using ResNets and benchmark image data sets, it is demonstrated that the proposed surrogate $F_\beta$ loss function is effective for optimizing $F_\beta$ scores under class imbalances in binary classification tasks compared with other loss functions. 

\vspace{1pc}

\end{abstract}

\section{Introduction}
\label{section:introduction}

An imbalanced data problem refers to the case that the distribution of class labels in the training data is not uniform but skewed \citep{He+Garcia_2008_KDE}. 
Such problem arises widely in real life applications of machine learning, such as computer vision \citep{Xiao_2010_CVPR_scene,Kubat_1998_ML_oil,Beijbom_2012_CVPR_reef}, biology \citep{Yu_2012_cancer}, medicine \citep{Grzymala-Busse_2004_Rough,Namee_2002_AIM_bias}, and fraud detection \citep{Chan_1998_KDD_fraud}. 
The imbalanced data problem can cause machine learning models, such as the convolutional neural networks, to perform poorly on test data, especially for minority classes \citep{Japkowicz_2002_IDA,He+Garcia_2008_KDE,VanHorn+Perona_2017_Arxiv,Buda_2018_NN_imbalance,Johnson_2019_survey}. 

In general, machine learning methods for addressing the imbalanced data problem can be grouped into two categories: data-level methods and algorithm-level methods \citep{Krawczyk_2016_open}. In data-level methods, the number of samples for each class is directly adjusted by over-sampling the minority classes, under-sampling the majority classes, or generating synthetic data for the minority classes. 
However, data-level methods often suffer from several problems. That is, over-sampling and synthetic data generation can introduce a large amount of duplicated samples, which can incur an overfitting problem and increased computational costs. And under-sampling can discard important samples, so that the test set performance of the models trained on partial data can be impaired.
Due to such potential problems, we focus on algorithm-level methods in this study.  
Algorithm-level methods are studied mostly on the cost-sensitive re-weighting methods, where loss functions are utilized to assign higher costs to the misclassified minority classes than to the misclassified majority classes.
Gradient-based learning algorithms, such as the stochastic gradient descent (SGD) method, search for the network parameters by minimizing the loss functions. 

On the other hand, the performance of a classifier is evaluated by performance measures such as the accuracy. In the context of imbalanced data, the accuracy has a severe limitation that it can lead to incorrect conclusions which favors the majority classes over the minority classes.
There are a number of performance measures that address this issue \citep{Johnson_2019_survey}. 
The $F_\beta$ score is one of the most widely used performance measures in imbalanced binary classification tasks. 
Note that the precision is defined as the proportion of positive predictions which are actually correct, 
and the recall is defined as the proportion of actual positives which are predicted correctly. 
The $F_\beta$ score is defined as a harmonic mean of precision and recall by
	\begin{equation}
	\label{eq:fbeta_score_primitive}
	F_\beta = 
	\frac{
		(1 + \beta^2) \cdot \text{Precision} \cdot \text{Recall}
	}{
		(\beta^2 \cdot \text{Precision}) + \text{Recall}
	}
	= 
	\left( 
	\lambda_\beta\cdot \text{Recall}^{-1} 
	+ (1 - \lambda_\beta)\cdot \text{Precision}^{-1}
	\right)^{-1}
	,
	\end{equation}
where the parameter $\beta > 0$ adjusts the relative weight,  $\lambda_\beta = \beta^2 / (1 + \beta^2)$, between precision and recall. 

However, most of the performance measures cannot be directly used as a loss function by gradient-based learning algorithms because they are usually not differentiable.
In most cases, loss functions which are used for optimizing neural network parameters are not identical to the performance measures, which results in sub-optimal model parameters in terms of performance measures. 
For instance, one of the most popular loss functions for binary classification is the binary cross-entropy (BCE) loss. 
It is well known that minimizing the BCE loss is equivalent to maximizing the likelihood for Bernoulli distribution, but a relationship between the BCE loss and performance measures has not been studied sufficiently. 

Our main contribution can be summarized as follows: (1) We investigate a relationship between classification performance measures and loss functions in terms of the gradients with respect to the model parameters. 
We derive explicit conditions that the gradients should satisfy approximately at critical points.
(2) We propose a differentiable surrogate loss function for optimization of $F_\beta$ score. 
(3) It is demonstrated that the gradient paths of the proposed loss function are approximately same to the gradient paths of the $F_\beta$ score through numerical experiments.
(4) We present a generalization of the proposed loss function for robustness to label noise. 

The rest of this paper is organized as follows.
In Section~\ref{section:related}, we provide the literature of related works. 
In Section~\ref{section:gradient}, the gradient conditions are analyzed for performance measures and standard loss functions. 
In Section~\ref{section:proposed}, we present the proposed surrogate $F_\beta$ loss function and its condition for gradients.
In Section~\ref{section:numerical}, numerical experiments are performed using ResNet models and benchmark  image data sets to compare the proposed loss function with other loss functions. 
Conclusion and future works are provided in Section~\ref{section:conclusion}.

\section{Related Work}
\label{section:related}

For binary classification of imbalanced data sets, weights are often set proportional to inverse class frequencies for cost-sensitive learning strategies \citep{Wang_2017_NIPS_tail}.
\cite{Cui_2019_CVPR} derived a novel formula for the effective number of samples and used it to propose a class-balanced loss function for cost-sensitive learning strategies. 
In this study, we analyze class-balanced loss functions to investigate their relationship with classification performance measures. 

In most cases, a machine learning model which yields the best value of a performance measure is desired, however, a learning algorithm is developed to optimize a loss function instead of the performance measure. 
Two major reasons that loss functions are preferred to performance measures by learning algorithms are their differentiability and convexity \citep{Bartlett_2006_JASA_convexity}. 
A direct extension of the $F_\beta$ score has been suggested for optimization of artificial neural networks \citep{Pastor_2013_soft_F_measure}, 
but it is a heuristically formulated loss function without mathematical justification.
There have been developed several algorithms to optimize differentiable lower bounds of performance measures \citep{Joachims_2005_ICML_svmperformance, Kar_2014_NIPS_online, Nara_2015_ICML_performance}, but the performance of deep neural networks trained with those lower bounds should be further improved \citep{Sanyal_2018_ML_optimizing}.
\cite{Eban_2017_ICAIS_scalable} proposed scalable optimization algorithms based on lower bounds for the performance measures, but its gradient path and performance were not analyzed sufficiently. 

In this study, we propose a surrogate loss function for $F_\beta$ score, and the same idea can be easily extended to define surrogate loss functions for other performance measures. 
We present mathematical conditions for the gradients of performance measures and the proposed loss function at critical points. 
The presented gradient conditions guarantee that the gradient paths of the proposed loss approximate those of the $F_\beta$ score asymptotically. 
Moreover, the proposed loss function can be used for scalable optimization by standard gradient-based optimization methods.

\section{Gradient Conditions of Performance Measures and Standard Loss Functions}
\label{section:gradient}

\subsection{Preliminaries}

We consider a $C$-class classification problem. Let $\calX$ denote the feature space and $\calY = \{0, 1, \ldots, C-1\}$ denote the label space. A sample is a set $\{(\bx_i, y_i)\}_{i=1}^n$ with each $(\bx_i, y_i) \in \calX \times \calY$. A classifier is a vector-valued function $\bff: \calX \rightarrow \bbR^C$, where each component function is denoted as $f_y$, $y\in \calY$. We assume that a classifier is a neural network with the softmax function as the output layer. 
A loss function $\calL$ is a real valued function, $\calL(\bff(\bx; \btheta), y)$, where $\btheta$ is a set of parameters. In the special case of binary classification problem, $C=2$, a classifier $\bff = (f_0, f_1)$ with the softmax output layer can be represented by a single real-valued function $f$ defined by $f \equiv f_1$ and $f_0 = 1-f$. In this case, a loss function can be denoted by $\calL(f(\bx; \btheta), y)$. 

A class-balanced loss function \citep{Cui_2019_CVPR} is defined by a loss function multiplied by a class-dependent weight, $w_y$, as 
\begin{equation}
	\calL_{\text{CB}}(\bff(\bx; \btheta), y) = w_y \calL(\bff(\bx; \btheta), y).
\end{equation}
In this study, we suppose that the class-dependent weight is set to the inverse class frequency, $w_y = n/n_y$ for $y\in\calY$, where $n_y$ is the number of training samples with label $y$ and $\sum_{y\in\calY} n_y = n$. 
%\cite{Cui_2019_CVPR} presented a generalized class-balanced loss function based on the concept of an effective number of samples. 

For classification tasks, most of the performance measures can be represented in terms of the entries of the confusion matrix. In the case of binary classification, the predicted class label $\hat{y}_i$ is defined by $\hat{y}_i = 1$ if $f (\bx_i; \btheta) \geq 0.5$ and $\hat{y}_i = 0$ otherwise. The entries of the confusion matrix can be written as 
\begin{equation}
\begin{split}
TP (\btheta) = \sum_{i=1}^n y_i \hat{y}_i, 
\quad 
TN (\btheta) = \sum_{i=1}^n (1-y_i) (1-\hat{y}_i), 
\\
FP (\btheta) = \sum_{i=1}^n (1-y_i) \hat{y}_i, 
\quad
FN (\btheta) = \sum_{i=1}^n y_i (1 - \hat{y}_i). 
\end{split}
\end{equation}
Let the sample proportions of actual positives, false negatives, and false positives be denoted by
\begin{equation}
\label{eq:sample_proportion}
	p_n = \frac{n_1}{n}, 
	\quad 
	q_{0n}= \frac{FN (\btheta)}{n_1}, 
	\quad
	q_{1n} = \frac{FP (\btheta)}{n - n_1}. 
\end{equation}
%Since the $n_1 = TP (\btheta) + FN (\btheta)$ and $n_0 = n - n_1 = TN (\btheta) + FP (\btheta)$, we can derive that $1-q_{0n} = TP (\btheta) / n_1$ and $1-q_{1n} = TN (\btheta) / (n-n_1)$. 
If we assume that the samples are independent and identically distributed, then the entries of the confusion matrix have the multinomial distribution with the total number of trials $n = TP (\btheta) + TN (\btheta) + FP (\btheta) + FN (\btheta)$. Hence, the expected values of the sample proportions can be defined by the true proportions as
\begin{equation}
	p = \bbE \left[ p_n \right] = \text{Pr} (y = 1), 
	\quad 
	q_0 = \bbE \left[ q_{0n} \right] = \text{Pr} (\hat{y} = 0 | y = 1), 
	\quad 
	q_1 = \bbE \left[ q_{1n} \right] = \text{Pr} (\hat{y} = 1 | y = 0). 
\end{equation} 
Based on the weak law of large numbers, we can derive that the sample proportions converge to the true proportions, i.e., $p_n \rightarrow p$, $q_{0n} \rightarrow q_0$, and $q_{1n} \rightarrow q_1$, as the sample sizes $n$, $n_1$, and $n-n_1$ tend to infinity.

In addition, for a real-valued function $g$ on $\calX$ and $y_0\in \{0, 1\}$, we define the conditional sample mean of $g(\bx)$ given $y=y_0$ by $\widehat{\bbE} \left[ g(\bx) | y=y_0 \right] = n_{y_0}^{-1} \sum_{i: y_i = y_0} g(\bx_i)$, which can be re-written as 
\begin{equation}
	\widehat{\bbE} \left[ g(\bx) | y=1 \right] = \frac{1}{n_1} \sum_{i = 1}^n y_i g(\bx_i), 
	\quad
	\widehat{\bbE} \left[ g(\bx) | y=0 \right] = \frac{1}{n_0} \sum_{i = 1}^n (1 - y_i) g(\bx_i)
	.
\end{equation}
Then, the sample proportions can be expressed as
$q_{0n} = \widehat{\bbE}\left[ 1 - \hat{y} | y=1 \right]$ and $q_{1n} = \widehat{\bbE} \left[ \hat{y} | y=0 \right]$.   
Recall that the predicted class label $\hat{y}$ is determined by the classifier output $f (\bx; \btheta)$, and it satisfies that $| \hat{y} - f ( \bx; \btheta) | \leq 0.5$. By replacing $\hat{y}$ with $f ( \bx; \btheta)$, we define the smoothed sample proportions by
\begin{equation}
\label{eq:smoothed_sample_proportion}
	\tilde{q}_{0n} = \widehat{\bbE}\left[ 1 - f ( \bx; \btheta) | y=1 \right], 
	\quad 
	\tilde{q}_{1n} = \widehat{\bbE} \left[ f ( \bx; \btheta) | y=0 \right]
	.
\end{equation}

\subsection{Gradient Condition of the $F_\beta$ Score}

Most of the performance measures, such as the accuracy, precision, $F_\beta$ score, and Jaccard score, can be represented as a combination of the entries of the confusion matrix \citep{Koyejo_2014_NIPS}. 
In this study, we focus on the $F_\beta$ score, which is defined in \eqref{eq:fbeta_score_primitive}. 
Note that the precision and the recall are defined by
\begin{equation}
	\text{Precision} = \frac{TP (\btheta)}{TP (\btheta) + FP (\btheta)}, 
	\quad  
	\text{Recall} = \frac{TP (\btheta)}{TP (\btheta) + FN (\btheta)}.
\end{equation}
By using the sample proportions in \eqref{eq:sample_proportion}, the $F_\beta$ score can be expressed as 
%the precision and the recall can be expressed as 
%\begin{equation}
%\label{eq:precision_analytic}
%\text{Precision} = \frac{TP (\btheta)}{TP (\btheta) + FP (\btheta)} = 
%\frac{n_1 (1-q_{0n})}{n_1 (1-q_{0n}) + (n-n_1) q_{1n}} = 
%\frac{p_n (1-q_{0n})}{p_n (1-q_{0n}) + (1 - p_n) q_{1n}}, 
%\end{equation}
%\begin{equation}
%\label{eq:recall_analytic}
%\text{Recall} = \frac{TP (\btheta)}{TP (\btheta) + FN (\btheta)} = 
%\frac{n_1 (1-q_{0n})}{n_1 (1-q_{0n}) + n_1 q_{0n}} = 
%1-q_{0n}.
%\end{equation}
%It follows that the $F_\beta$ score can be expressed as 
\begin{equation}
\label{eq:fbeta_score_analytic}
F_\beta = 
\frac{
    (1 + \beta^2) \cdot p_n (1 - q_{0n})
}{
    p_n (\beta^2 + 1 - q_{0n} - q_{1n}) + q_{1n}
}
.
\end{equation}
Even though the $q_{0n}$ and $q_{1n}$ are functions of the parameters $\btheta$, they cannot be differentiated with respect to the parameters because the entries of the contingency matrix are not smooth functions of $\btheta$. 
On the other hand, the limit of the $F_\beta$ score as $n \rightarrow \infty$ can be expressed in terms of the true proportions as 
\begin{equation}
\lim_{n\rightarrow \infty} F_\beta = 
\tilde{F}_\beta =
\frac{
    (1 + \beta^2) \cdot p (1 - q_0)
}{
    p (\beta^2 + 1 - q_0 - q_1) + q_1
}. 
\end{equation}
Since the true proportions, $q_0$ and $q_1$, are smooth functions of $\btheta$, the limit of $F_\beta$ is a smooth function. 
Considering a gradient-based learning algorithms, taking the gradient of the logarithm of the limit function yields
\begin{equation}
\frac{\partial }{\partial \btheta} \log \tilde{F}_\beta = z^{-1} \cdot 
\left(  -q_0' (\beta^2 \cdot p -pq_1 +q_1) + q_1' (p - pq_0 - 1 + q_0)
\right) 
,
\end{equation}
where $q_0' = \nabla_\btheta q_0$ and $q_1' = \nabla_\btheta q_1$ are the gradients of the true proportions, and $z = (1 - q_0)\cdot (p (\beta^2 + 1 - q_0 - q_1) + q_1)$. 
From the condition at the critical point, $\partial \log \tilde{F}_\beta / \partial \btheta = {\bf 0}$, we obtain the following condition of the gradients: 
\begin{equation}
\label{eq:gradient_fbeta}
	-\frac{q_0'}{1 - q_0} = \frac{q_1'}{\beta^2 \cdot p/(1-p) + q_1}. 
\end{equation}

\subsection{Standard Loss Functions}

We consider the following two standard loss functions for binary classification:
\begin{itemize}
\item 
The binary cross-entropy (BCE) loss can be expressed as
\begin{equation}
	\label{eq:bce_loss}
	\calL_\text{BCE} ( \bff(\bx; \btheta), y) = - \log f_y (\bx; \btheta) = - y \log f (\bx; \btheta) - (1-y) \log ( 1 - f(\bx; \btheta) ). 
\end{equation}

\item 
The mean absolute error (MAE) loss, also called as the $l_1$ loss, can be written as
\begin{equation}
	\label{eq:mae_loss}
	\calL_\text{MAE} ( \bff(\bx; \btheta), y) = \| \be_y - \bff(\bx; \btheta) \|_1 = 2 - 2 f_y (\bx; \btheta)
	= 2y (1 - f(\bx; \btheta)) + 2(1-y) f(\bx; \btheta), 
\end{equation}
where $\be_y$ is the vector whose $y$th element is 1 and the other elements are zeros. 

%\item
%More generally, \cite{Zhang_2018_NEURIPS_generalized} proposed a generalized cross entropy loss, which extends the BCE and MAE losses. It can be expressed as 
%\begin{equation}
%	\calL_q ( \bff(\bx; \btheta), y) = \frac{(1 - f_y (\bx; \btheta)^q)}{q} 
%	= \frac{y (1 - f (\bx; \btheta)^q)}{q} + \frac{(1-y) (1 - (1 - f (\bx; \btheta))^q)}{q}, 
%\end{equation}
%where $q\in (0, 1]$. It can be shown that $\lim_{q\rightarrow 0} \calL_q ( \bff(\bx; \btheta), y) = \calL_\text{BCE} ( \bff(\bx; \btheta), y)$ and $\calL_0 ( \bff(\bx; \btheta), y) = 0.5 \times \calL_\text{MAE} ( \bff(\bx; \btheta), y)$.
\end{itemize}

\subsection{Gradient Condition of the BCE Loss}

With the class weight $w_y = n/n_y$, the sample mean of the class-balanced BCE losses can be expressed as 
\begin{equation}
\label{eq:bce_cbloss}
\begin{split}
	\widehat{\calR}^w_\text{BCE}(\btheta) 
	& 
	= 
	\frac{1}{n} \sum_{i=1}^n w_{y_i} \calL_\text{BCE} (\bff(\bx_i; \btheta), y_i) 
	\\
	& 
	= -  \frac{1}{n} \sum_{i=1}^n \left( 
		\frac{n y_i}{n_1} \log f (\bx_i; \btheta) + \frac{n (1-y_i)}{n-n_1} \log ( 1 - f(\bx_i; \btheta) )
	\right)
	\\
	& 
	= \widehat{\bbE} \left[ \left. - \log f (\bx; \btheta) \right| y=1 \right]
	+ \widehat{\bbE} \left[ \left. - \log (1 - f (\bx; \btheta)) \right| y=0 \right]
	\\
	& 
	\geq 
	- \log \widehat{\bbE} \left[ \left. f (\bx; \btheta) \right| y=1 \right]
	- \log \widehat{\bbE} \left[ \left. (1 - f (\bx; \btheta)) \right| y=0 \right]
	,
\end{split}
\end{equation}
where the inequality at the last line follows from the Jensen's inequality. 
After all, from \eqref{eq:smoothed_sample_proportion} and \eqref{eq:bce_cbloss}, the sample mean of the class-balanced BCE losses can be approximately represented by 
\begin{equation}
\label{eq:bce_cbloss_2}
	\widehat{\calR}^w_\text{BCE}(\btheta) 
	\approx 
	- \log ( 1- \tilde{q}_{0n} )
	- \log ( 1 - \tilde{q}_{1n} ), 
\end{equation}
where the approximation error is determined by the Jensen's inequality in \eqref{eq:bce_cbloss}. 
Note that, unlike the sample proportions $q_{0n}$ and $q_{1n}$, the smoothed sample proportions $\tilde{q}_{0n}$ and $\tilde{q}_{1n}$ are differentiable with respect to the parameters $\btheta$. 
From $\partial \widehat{\calR}^w_\text{BCE}(\btheta) / \partial \btheta = {\bf 0}$, we can obtain the following condition for the gradients:
\begin{equation}
\label{eq:gradient_bce}
	- \frac{\tilde{q}_{0n}'}{1 - \tilde{q}_{0n}}
	= 
	\frac{\tilde{q}_{1n}'}{1 - \tilde{q}_{1n}}
	,
\end{equation}
where $\tilde{q}_{0n}' = \nabla_\btheta \tilde{q}_{0n}$ and $\tilde{q}_{1n}' = \nabla_\btheta \tilde{q}_{1n}$.

\subsection{Gradient Condition of the MAE Loss}

The sample mean of the class-balanced MAE losses can be approximately represented by 
\begin{equation}
\label{eq:mae_cbloss}
\begin{split}
	\widehat{\calR}^w_\text{MAE}(\btheta) 
	&
	=
	\frac{1}{n} \sum_{i=1}^n w_{y_i} \calL_\text{MAE} (\bff(\bx_i; \btheta), y_i) 
	\\
	& 
	= \frac{2}{n} \sum_{i=1}^n \left( 
	   \frac{n y_i}{n_1} (1 - f (\bx_i; \btheta)) + \frac{n (1-y_i)}{n-n_1} f(\bx_i; \btheta)
	\right)
	\\
	& 
	= 2 \widehat{\bbE} \left[ \left. 1 - f (\bx; \btheta) \right| y=1 \right]
	+ 2 \widehat{\bbE} \left[ \left. f (\bx; \btheta) \right| y=0 \right]
	\\
	& 
	= 2 \tilde{q}_{0n} + 2 \tilde{q}_{1n}
	.
\end{split}
\end{equation}
From $\partial \widehat{\calR}^w_\text{MAE}(\btheta) / \partial \btheta = {\bf 0}$, we can obtain the following condition for the gradients:
\begin{equation}
\label{eq:gradient_mae}
	- \tilde{q}_{0n}'
	= 
	\tilde{q}_{1n}'
	.
\end{equation}

\section{Proposed Surrogate Loss for the $F_\beta$ Score}
\label{section:proposed}

Motivated from the gradient conditions of the BCE loss and the MAE loss in \eqref{eq:gradient_bce} and \eqref{eq:gradient_mae}, 
we propose the surrogate loss function for the $F_\beta$ score, which is defined by 
\begin{equation}
\label{eq:surrogate_fbeta_loss}
	\calL_{F_\beta} (\bff(\bx; \btheta), y)
	= - y \log f (\bx; \btheta) + (1-y) \log \left( \beta^2 \cdot \frac{p}{1-p} + f(\bx; \btheta) \right)
	, 
\end{equation}
where $\beta > 0$ controls the balance between the precision and recall as in the $F_\beta$ score, and $0<p<1$ represents the proportion of the positive samples, $p = n_1 / n$. 
Note that the positive part ($y = 1$) of the surrogate loss function is equal to that of the BCE loss, but the negative part ($y = 0$) is different. The reason can be found by analyzing the gradient condition of the surrogate loss function as follows. The sample mean of the class-balanced surrogate loss function is 
\begin{equation}
\label{eq:surrogate_cbloss}
\begin{split}
	\widehat{\calR}^w_{F_\beta}(\btheta) 
	&
	=
	\frac{1}{n} \sum_{i=1}^n w_{y_i} \calL_{F_\beta} (\bff(\bx_i; \btheta), y_i) 
	\\
	& 
	= \frac{1}{n} \sum_{i=1}^n \left( 
	   - \frac{n y_i}{n_1} \log f (\bx_i; \btheta) 
	   + \frac{n (1-y_i)}{n-n_1} \log \left( \beta^2 \cdot \frac{p}{1-p} + f(\bx_i; \btheta)  \right)
	\right)
	\\
	& 
	= \widehat{\bbE} \left[ \left. - \log f (\bx; \btheta) \right| y=1 \right]
	+ \widehat{\bbE} \left[ \left. \log \left( \beta^2 \cdot \frac{p}{1-p} + f(\bx_i; \btheta)  \right) \right| y=0 \right]
	\\
	& 
	\approx -\log (1 - \tilde{q}_{0n}) + \log \left( \beta^2 \cdot \frac{p}{1-p} + \tilde{q}_{1n}  \right)
	.
\end{split}
\end{equation}
From $\partial \widehat{\calR}^w_{F_\beta}(\btheta) / \partial \btheta = {\bf 0}$, we can derive that
\begin{equation}
\label{eq:gradient_surrogate_fbeta}
	- \frac{\tilde{q}_{0n}'}{1 - \tilde{q}_{0n}}
	= 
	\frac{\tilde{q}_{1n}'}{\beta^2 \cdot p / (1-p) + \tilde{q}_{1n}}
	.
\end{equation}
Note that the gradient property for the $F_\beta$ score in \eqref{eq:gradient_fbeta} is equivalent to that for the surrogate loss in \eqref{eq:gradient_surrogate_fbeta} except that one employs the true proportions and the other employs the smoothed sample proportions. Hence, the additional parameters $\beta$ and $p$ for the surrogate loss possess the same roles as in the $F_\beta$ score, and they can be utilized to control the $F_\beta$ score of the trained neural network models. 

\begin{figure}
\centering
\begin{tabular}{cc}
\includegraphics[width=6cm]{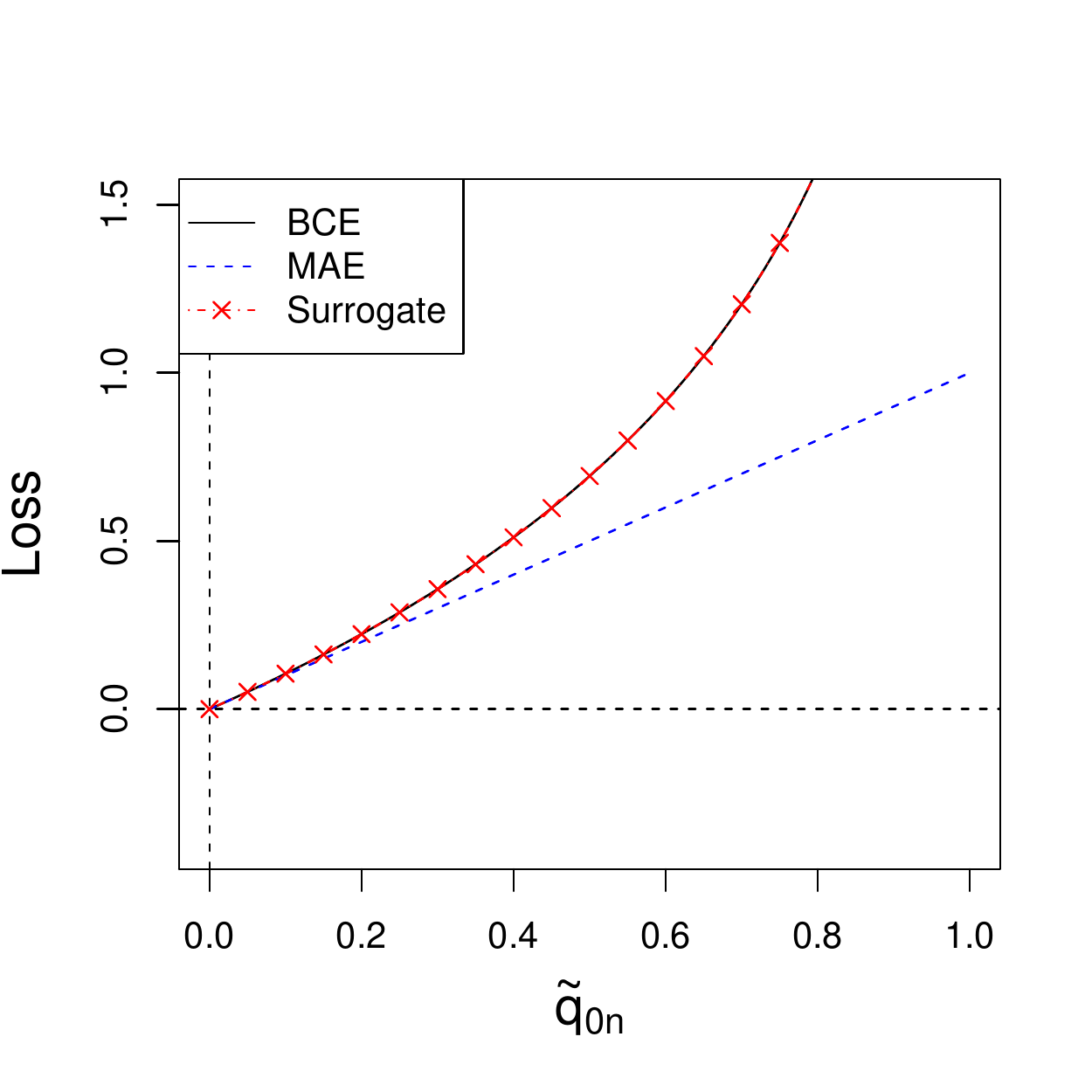} &
\includegraphics[width=6cm]{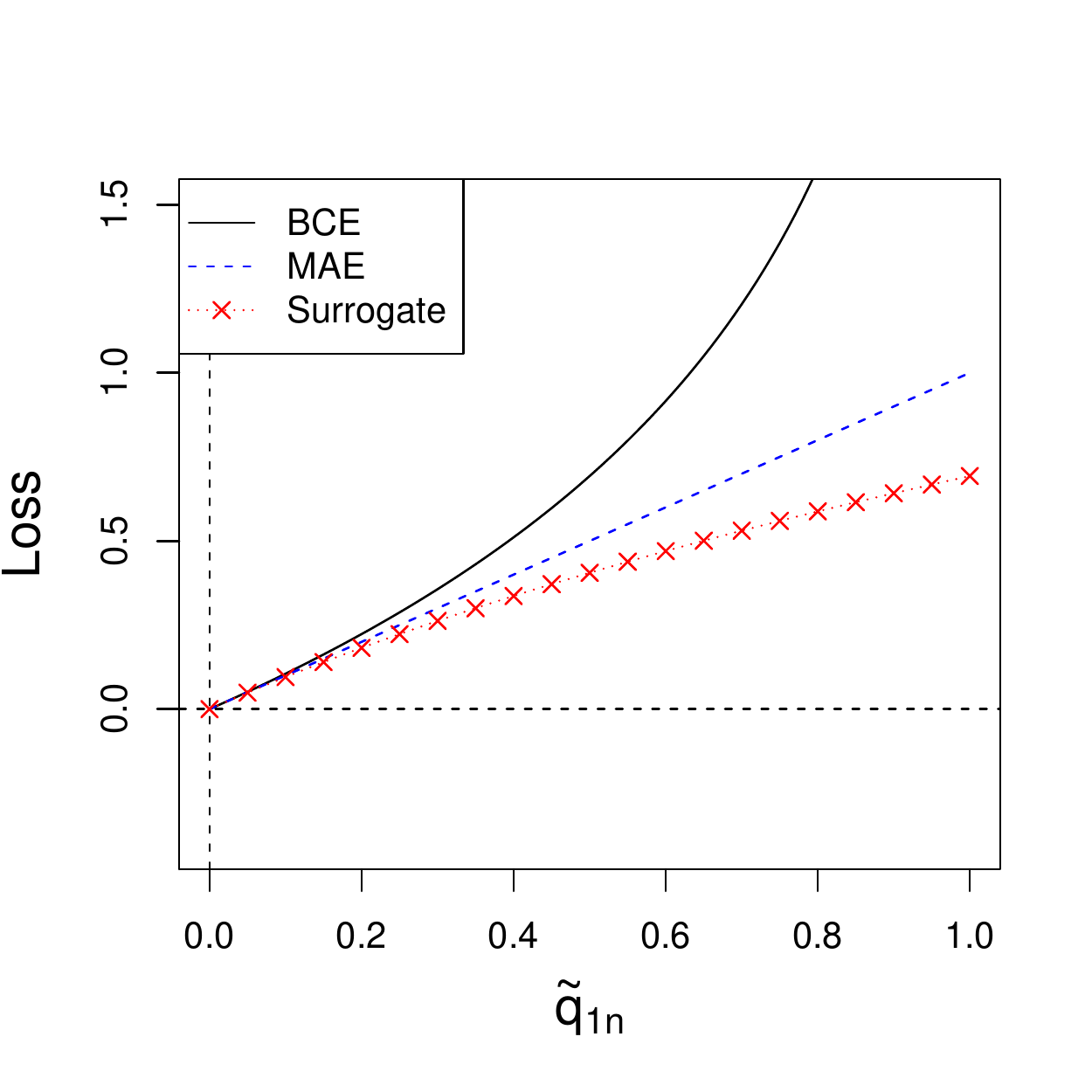} \\
(a) & 
(b)
\end{tabular}
\caption{\label{fig:compare_losses_diagram}
Illustration of (a) the functions of $\tilde{q}_{0n}$ and (b) the functions of $\tilde{q}_{1n}$ comprising of the the expressions in \eqref{eq:bce_cbloss_2}, \eqref{eq:mae_cbloss}, and \eqref{eq:surrogate_cbloss}
for the approximate sample mean of the loss functions.
}
\end{figure}

From the expressions in \eqref{eq:bce_cbloss_2}, \eqref{eq:mae_cbloss}, and \eqref{eq:surrogate_cbloss}, 
we can find that every expression for the sample mean of the loss functions consists of two terms: a function of $\tilde{q}_{0n}$ and a function of $\tilde{q}_{1n}$. 
Figure~\ref{fig:compare_losses_diagram}(a) illustrates the first term, i.e., the function of $\tilde{q}_{0n}$, for each of the BCE, MAE, and surrogate $F_\beta$ loss functions. 
In specific, the functions corresponding to the first term can be written as $z=-\log (1-q)$, $z=q$, and $z=-\log(1 - q)$ with $q = \tilde{q}_{0n}$.
Likewise, Figure~\ref{fig:compare_losses_diagram}(b) compares the second term, i.e., the function of $\tilde{q}_{10}$, for the BCE, MAE, and surrogate $F_\beta$ loss functions. 
The three functions corresponding to the second term can be written as 
$-\log (1-q)$, $q$, and $\log(1 + q)$ with $q = \tilde{q}_{1n}$, respectively, when $\beta=1$ and $p=0.5$.
Note that the first term of the surrogate $F_\beta$ loss is equal to the first term of the BCE loss, but the second term of the surrogate $F_\beta$ loss is quite unique and different from those of the other losses.

\section{Numerical Experiments}
\label{section:numerical}

In this section, we conduct numerical experiments for evaluating the gradient paths and the performances of the proposed surrogate $F_\beta$ loss function. We compare several standard loss functions listed as follows:
\begin{enumerate}
	\item[(a)] The macro soft $F_\beta$ loss is the direct extension of the $F_\beta$ score \citep{Pastor_2013_soft_F_measure}, which can be written as 
			\begin{equation}
			\calL (\bff(\bx; \btheta), y)
			= 1 - \tilde{F}_\beta, 
			\end{equation}
		where $\tilde{F}_\beta$ is the smoothed $F_\beta$ score defined by
			\begin{equation}
			\tilde{F}_\beta = 
			\frac{
			    (1 + \beta^2) \cdot p_n (1 - \tilde{q}_{0n})
			}{
			    p_n (\beta^2 + 1 - \tilde{q}_{0n} - \tilde{q}_{1n}) + \tilde{q}_{1n}
			}
			, 
			\end{equation}
		and the $\tilde{q}_{0n}$ and $\tilde{q}_{1n}$ are the smoothed sample proportions defined in 
		\eqref{eq:smoothed_sample_proportion}.
	\item[(b)] The BCE loss in \eqref{eq:bce_loss}.
	\item[(c)] The MAE loss in \eqref{eq:mae_loss}.
\end{enumerate}

For the experiments, ResNet models \citep{He_CVPR_2016_ResNet} were trained and tested on the CIFAR-10\footnote{https://www.cs.toronto.edu/~kriz/cifar.html} and Fashion-MNIST\footnote{https://github.com/zalandoresearch/fashion-mnist} data sets. 
The experiments were implemented based on PyTorch \citep{Paszke_2019_PyTorch}
running on Ubuntu 18 with Intel(R) Xeon CPU of 2.30GHz and GeForce RTX 2080 Ti.
See the following section for the detailed experimental settings.

\subsection{Experimental Settings} 

ResNet-18 was used for binary classification of T-shirt/top (the first class) and the other classes in the Fashion-MNIST data set. ResNet-34 was used for binary classification of airplane images (the first class) and the other classes in the CIFAR-10 data set. 
In each of the data sets, the first class accounts for 10 percent of the whole data set. ResNet models employed ReLU in hidden layers and softmax layer at output layer. 
For training ResNet models, we used a mini-batch size of 100 and stochastic gradient descent (SGD) with 0.9 momentum, a weight decay of $10^{-4}$, and learning rate of 0.01. Each experiment consisted of 120 epochs in total. We repeated the experiments five times independently. In each experiment, network weights were randomly initialized and 10 percent of the training set was randomly separated into validation set and the rest were remained as training set and used for training ResNet models. For data augmentation and preprocessing, each image was transformed by $32\times 32$ random crop after padding 4 pixels on each side, random horizontal flip, and normalization by per-pixel mean subtraction. Since ResNet takes three channels as input but the images in the Fashion-MNIST are gray scale single channel images, each image was repeated into three channels.

\subsection{Results on Fashion-MNIST Data Set}

\begin{figure}
\centering
\begin{tabular}{cc}
\includegraphics[height=4.5cm]{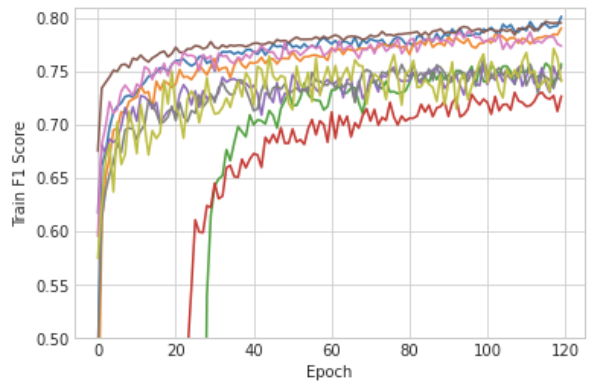} &
\includegraphics[height=4.5cm]{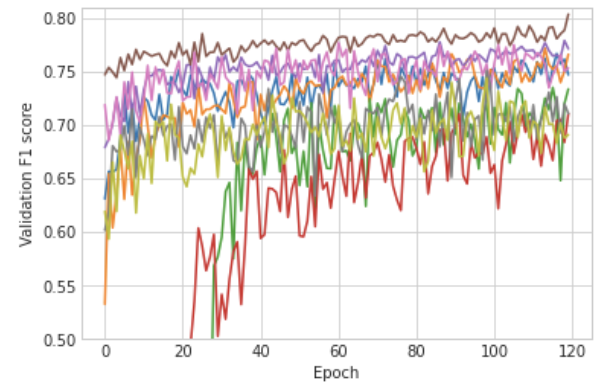}\\
(a) F1 score on training set &
(b) F1 score on validation set \\
\multicolumn{2}{c}{\includegraphics[height=4.5cm]{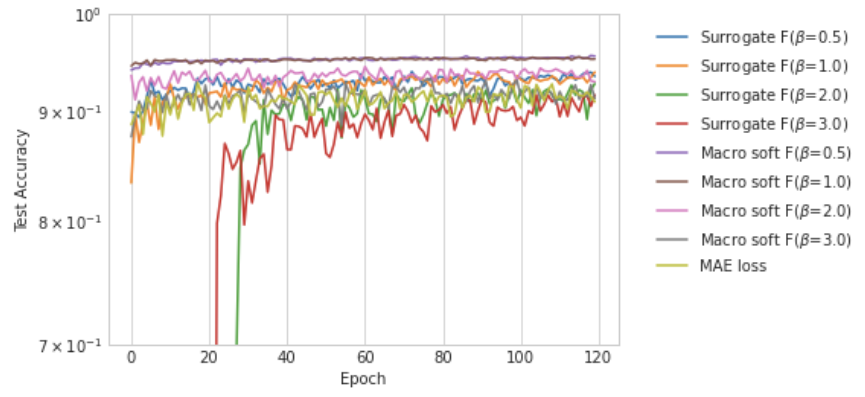}} \\
\multicolumn{2}{c}{(c) Accuracy on test set}
\end{tabular}
\caption{\label{fig:convergence_fashionmnist}The convergence of the ResNet-18 models trained by using the proposed loss functions and the other loss functions on the Fashion-MNIST data set. Model performances are measured by 
(a) F1 score on the training set, (b) F1 score on the validation set, and (c) accuracy on the test set. The performance values are the median from the five repeated experiments.
}
\end{figure}

For the Fashion-MNIST data set, the ResNet-18 models were trained by the proposed surrogate $F_\beta$ losses and the other standard loss functions. 
Figure~\ref{fig:convergence_fashionmnist} illustrates the convergence of the ResNet-18 models trained by using the proposed loss functions and the other loss functions, where all the loss functions were class-balanced by the inverse class frequency. 
We present three performance measures, which are the F1 score on the training set, F1 score on the validation set, and accuracy on the test set. Each performance values are the median of the five repeated experiments. 
We removed the BCE loss function because the models trained by the BCE loss function failed to achieve any improvement in performance, maybe because the class-balanced BCE loss requires a larger number of epochs greater than 120 for training ResNets. 
We note that the surrogate $F_\beta$ loss with a large $\beta$ value, i.e., $\beta=3$ or $\beta=2$, requires a relatively larger number of epoch for convergence, which is partly due to the relatively small increment rate of the loss function, which is illustrated in Figure~\ref{fig:compare_losses_diagram}(b). 

\begin{figure}
\centering
\begin{tabular}{cc}
\includegraphics[height=4.5cm]{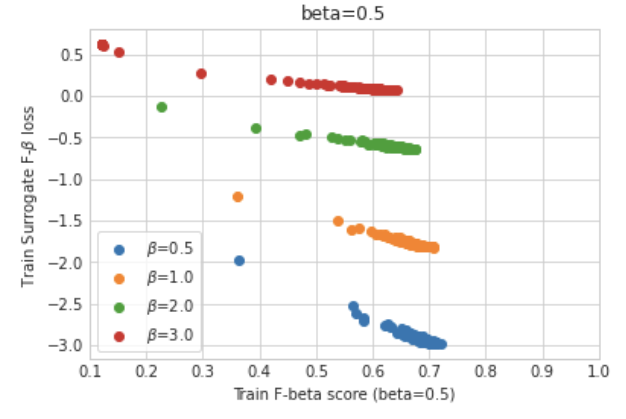} &
\includegraphics[height=4.5cm]{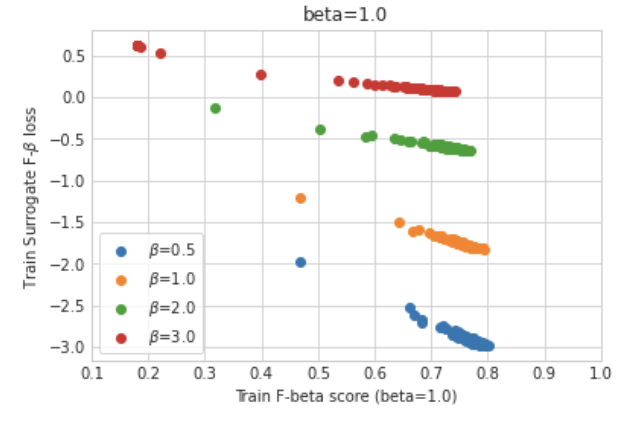}\\
(a) Surrogate $F_\beta$ loss v.s. $F_{0.5}$ score &
(b) Surrogate $F_\beta$ loss v.s. $F_{1.0}$ score \\
\includegraphics[height=4.5cm]{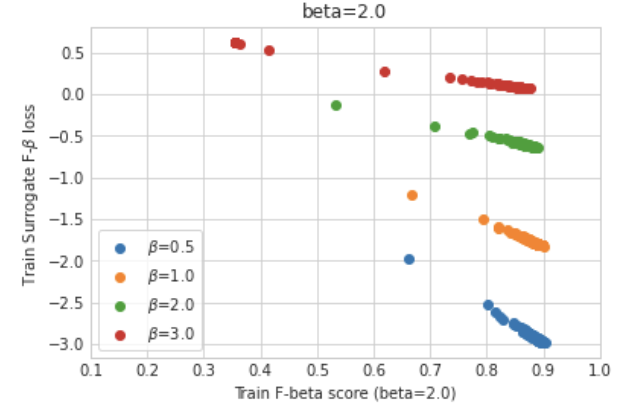} &
\includegraphics[height=4.5cm]{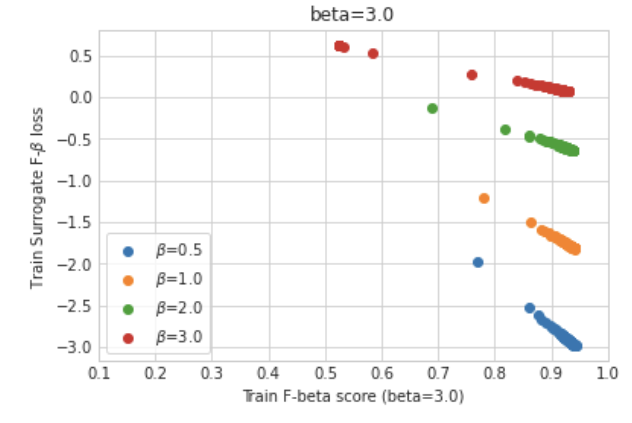}\\
(c) Surrogate $F_\beta$ loss v.s. $F_{2.0}$ score &
(d) Surrogate $F_\beta$ loss v.s. $F_{3.0}$ score
\end{tabular}
\caption{\label{fig:path_fashionmnist_surrogate}Scatter plot between the surrogate $F_\beta$ loss and the $F_\beta$ score on the training set of the Fashion-MNIST data set at the first repetition of the experiment. 
The $\beta$ value for the $F_\beta$ score were set to (a) 0.5, (b) 1.0, (c) 2.0, and (d) 3.0. 
}
\end{figure}

Figure~\ref{fig:path_fashionmnist_surrogate} shows the scatter plot between the surrogate $F_\beta$ loss and the $F_\beta$ score on the training set at the first repetition of the experiment, which compares the gradient paths of the surrogate $F_\beta$ loss and the those of the $F_\beta$ score. 
It is clear that the surrogate $F_\beta$ loss is linearly correlated with the $F_\beta$ score over all choices of the $\beta$ values. 
It implies that, even if the gradient paths of the surrogate $F_\beta$ loss function may fluctuate over epochs (see, e.g., Figure~\ref{fig:convergence_fashionmnist}(a)), they are perfectly aligned with the gradient paths of the actual $F_\beta$ score. 
The figure demonstrates that the surrogate $F_\beta$ loss can be an effective method for optimization of the $F_\beta$ score. 

\begin{figure}
\centering
\begin{tabular}{cc}
\includegraphics[height=4.5cm]{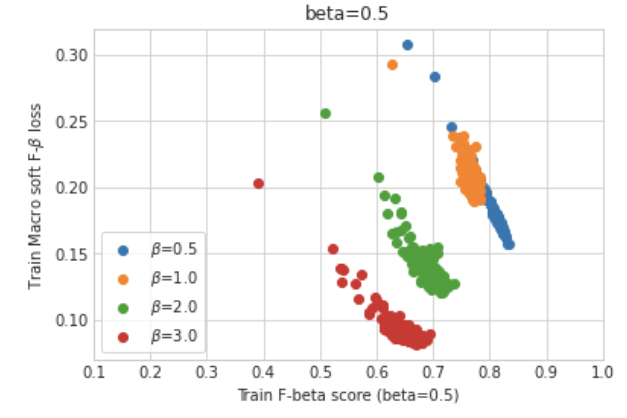} &
\includegraphics[height=4.5cm]{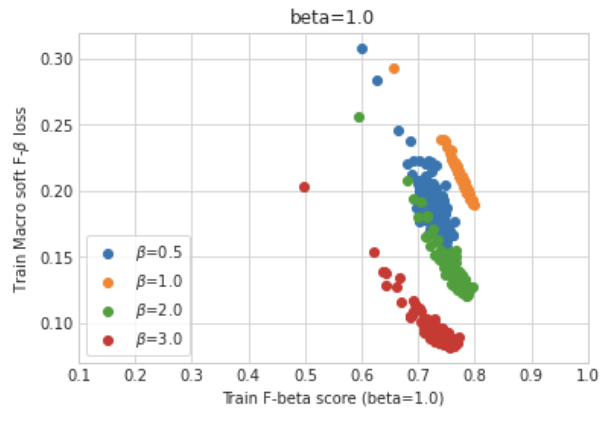}\\
(a) Macro soft $F_\beta$ loss v.s. $F_{0.5}$ score &
(b) Macro soft $F_\beta$ loss v.s. $F_{1.0}$ score \\
\includegraphics[height=4.5cm]{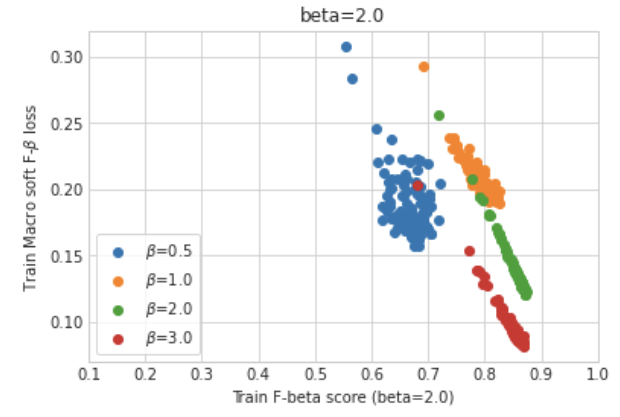} &
\includegraphics[height=4.5cm]{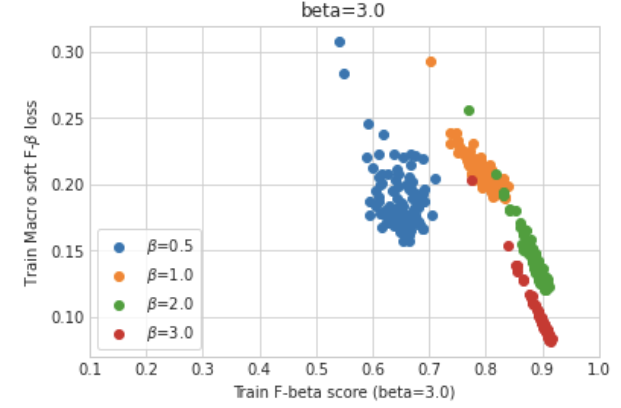}\\
(c) Macro soft $F_\beta$ loss v.s. $F_{2.0}$ score &
(d) Macro soft $F_\beta$ loss v.s. $F_{3.0}$ score
\end{tabular}
\caption{\label{fig:path_fashionmnist_macrosoft}Scatter plot between the macro soft $F_\beta$ loss and the $F_\beta$ score on the training set of the Fashion-MNIST data set at the first repetition of the experiment. 
The $\beta$ value for the $F_\beta$ score were set to (a) 0.5, (b) 1.0, (c) 2.0, and (d) 3.0. 
}
\end{figure}

For comparison with the macro soft $F_\beta$ loss, Figure~\ref{fig:path_fashionmnist_macrosoft} shows the scatter plot between the macro soft $F_\beta$ loss and the $F_\beta$ score on the training set at the first repetition of the experiment. 
In this figure, we can find that the macro soft $F_\beta$ loss is not linearly correlated with the $F_\beta$ score except a few cases of $\beta$ values.
It implies that optimization of ResNet models by using the macro soft $F_\beta$ loss may not yield optimal $F_\beta$ scores. 

\subsection{Results on CIFAR-10 Data Set}

\begin{figure}
\centering
\begin{tabular}{cc}
\includegraphics[height=4.5cm]{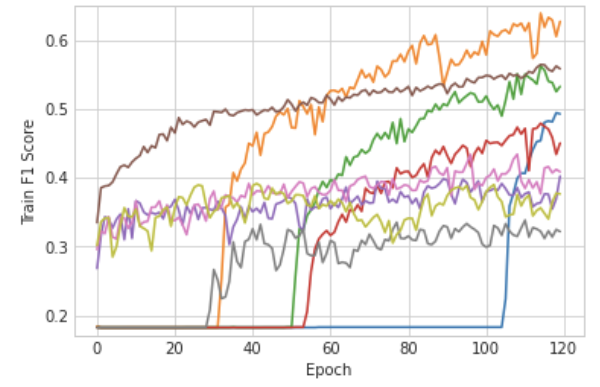} &
\includegraphics[height=4.5cm]{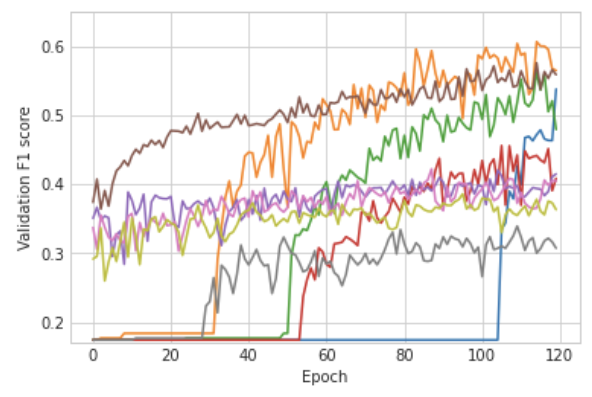}\\
(a) F1 score on training set &
(b) F1 score on validation set \\
\multicolumn{2}{c}{\includegraphics[height=4.5cm]{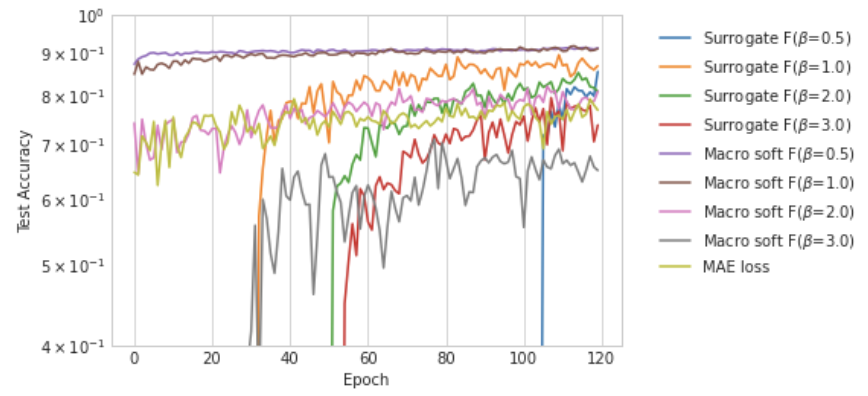}} \\
\multicolumn{2}{c}{(c) Accuracy on test set}
\end{tabular}
\caption{\label{fig:convergence_cifar10}The convergence of the ResNet-34 models trained by using the proposed loss functions and the other loss functions. Model performances are measured by 
(a) F1 score on the training set, (b) F1 score on the validation set, and (c) accuracy on the test set. The performance values are the median from the five repeated experiments.
}
\end{figure}

ResNet-34 is considered a deeper model with more parameters than ResNet-18.
Figure~\ref{fig:convergence_cifar10} illustrates the convergence of the ResNet-34 models trained by using the proposed loss functions and the other loss functions, where all the loss functions were class-balanced by the inverse class frequency. 
In Figure~\ref{fig:convergence_cifar10}, the ResNet models trained by using the surrogate $F_\beta$ loss functions converge relatively fast than the other loss functions, even if they require more epochs to find good initial weights. 
On the other hand, other loss functions except for the macro soft $F_1$ loss converge relatively slowly with respect to the $F_1$ score, 
which is partly due to the mismatch between loss function and performance measure. 
Moreover, Figure~\ref{fig:convergence_cifar10}(c) shows that the macro soft $F_\beta$ loss achieves the best accuracy on test set, but
the accuracy of 0.9 is the base accuracy value for the imbalanced data set in this binary classification task. 
Instead, in Figure~\ref{fig:convergence_cifar10}(b), we can see that the surrogate $F_\beta$ loss achieves the best $F_1$ score on validation set, 
which implies that the accuracy measure can be largely misleading. 

\begin{figure}
\centering
\begin{tabular}{cc}
\includegraphics[height=4.5cm]{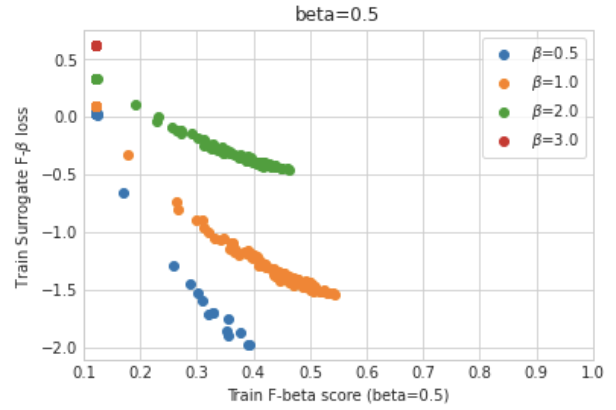} &
\includegraphics[height=4.5cm]{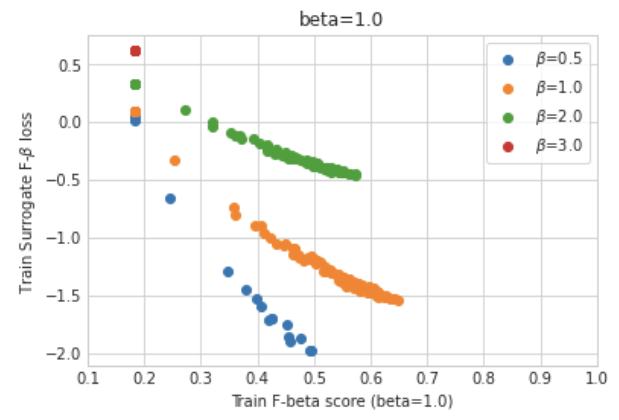}\\
(a) Surrogate $F_\beta$ loss v.s. $F_{0.5}$ score &
(b) Surrogate $F_\beta$ loss v.s. $F_{1.0}$ score \\
\includegraphics[height=4.5cm]{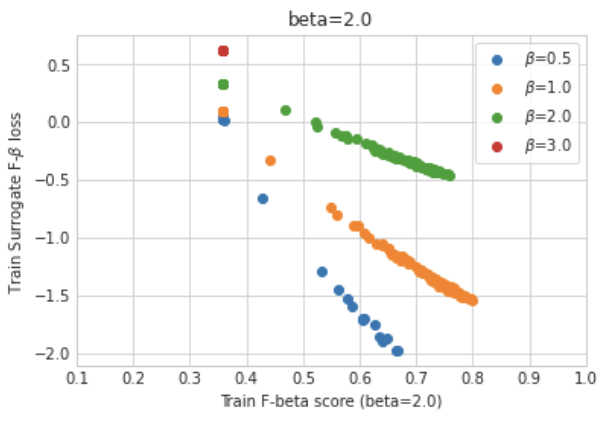} &
\includegraphics[height=4.5cm]{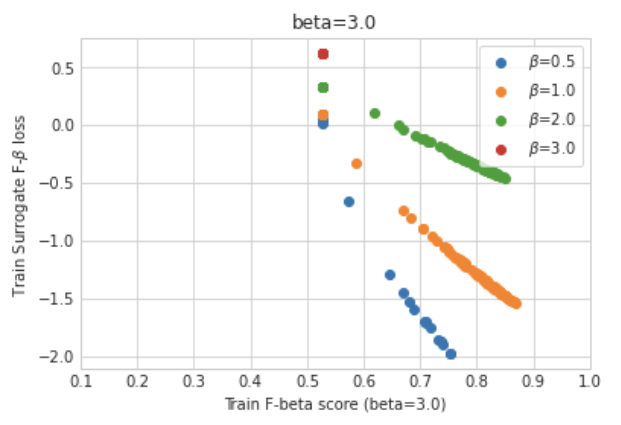}\\
(c) Surrogate $F_\beta$ loss v.s. $F_{2.0}$ score &
(d) Surrogate $F_\beta$ loss v.s. $F_{3.0}$ score
\end{tabular}
\caption{\label{fig:path_cifar10_surrogate}Scatter plot between the surrogate $F_\beta$ loss and the $F_\beta$ score on the training set at the first repetition of the experiment. 
The $\beta$ value for the $F_\beta$ score were set to (a) 0.5, (b) 1.0, (c) 2.0, and (d) 3.0. 
}
\end{figure}

\begin{figure}
\centering
\begin{tabular}{cc}
\includegraphics[height=4.5cm]{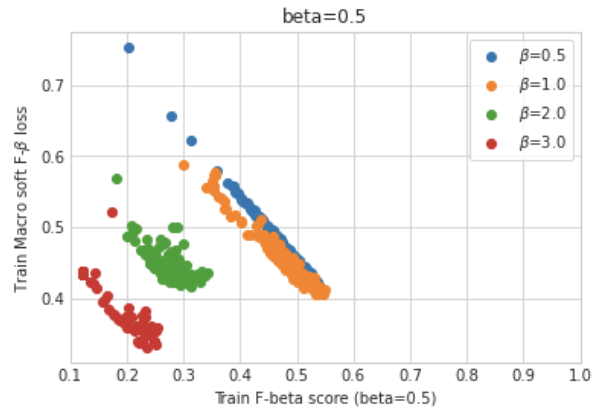} &
\includegraphics[height=4.5cm]{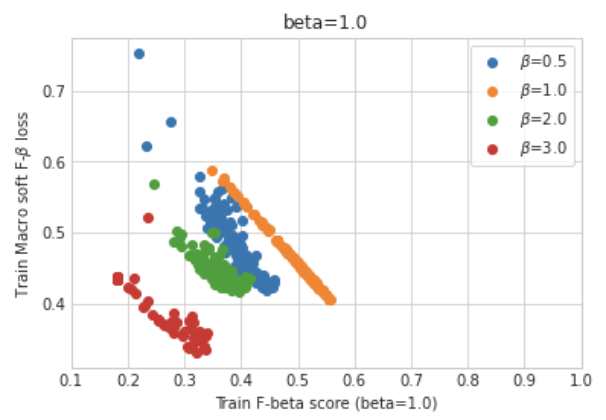}\\
(a) Macro soft $F_\beta$ loss v.s. $F_{0.5}$ score &
(b) Macro soft $F_\beta$ loss v.s. $F_{1.0}$ score \\
\includegraphics[height=4.5cm]{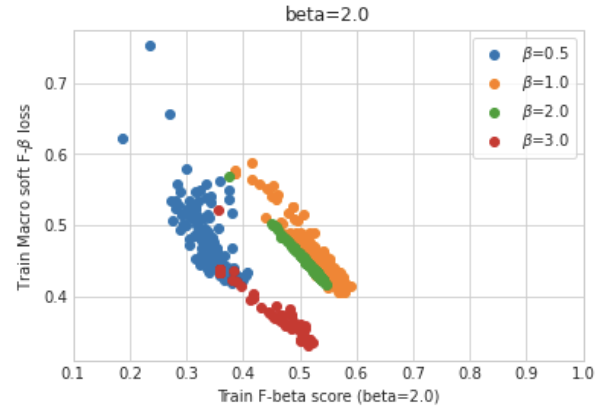} &
\includegraphics[height=4.5cm]{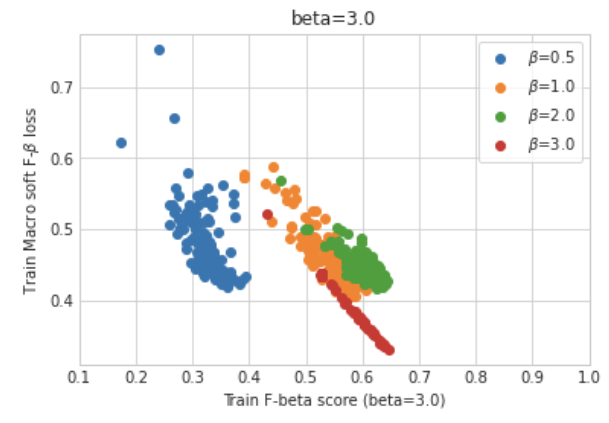}\\
(c) Macro soft $F_\beta$ loss v.s. $F_{2.0}$ score &
(d) Macro soft $F_\beta$ loss v.s. $F_{3.0}$ score
\end{tabular}
\caption{\label{fig:path_cifar10_macrosoft}Scatter plot between the macro soft $F_\beta$ loss and the $F_\beta$ score on the training set at the first repetition of the experiment. 
The $\beta$ value for the $F_\beta$ score were set to (a) 0.5, (b) 1.0, (c) 2.0, and (d) 3.0. 
}
\end{figure}

Figure~\ref{fig:path_cifar10_surrogate} shows the scatter plot between the surrogate $F_\beta$ loss and the $F_\beta$ score on the training set at the first repetition of the experiment, which compares the gradient paths of the surrogate $F_\beta$ loss and those of the $F_\beta$ score. 
It is clear that the surrogate $F_\beta$ loss is linearly correlated with the $F_\beta$ score over all choices of the $\beta$ values. 
Figure~\ref{fig:path_cifar10_macrosoft}, on the other hand, shows the scatter plot between the macro soft $F_\beta$ loss and the $F_\beta$ score. 
After comparing the figures, it is clear that the surrogate $F_\beta$ loss function can generate gradient paths which effectively approximates the gradient paths of the actual $F_\beta$ score.

\section{Conclusion and Extensions}
\label{section:conclusion}

In this study, we investigated the conditions of the gradients for the $F_\beta$ score, which is one of the most popular performance measure in imbalanced data classification. 
The derived gradient conditions were compared with those of the standard loss functions, 
and we proposed a loss function which is surrogate to the $F_\beta$ score. 
The proposed loss function, which is called the surrogate $F_\beta$ loss, can effectively approximate the gradient paths of the $F_\beta$ score when the network parameters are optimized by gradient-based learning algorithms such as the stochastic gradient descent method. 
The experimental results demonstrated that the proposed surrogate $F_\beta$ loss can be used to effectively optimize $F_\beta$ score of deep neural networks such as ResNet models. 

The approach adopted in this study for analyzing the gradient conditions of the $F_\beta$ score can be extended to other performance measures which are computed based on contingency matrix to build surrogate loss functions. 
The proposed surrogate loss function can be used for a scalable optimization method where mini-batch or parallel optimization approaches are adopted. 

Moreover, the proposed loss function can be extended to a generalized cross entropy to cope with noisy labels in multiclass classification problems \citep{Zhang_2018_NEURIPS_generalized}. 
For instance, the generalized surrogate $F_\beta$ loss can be written as
	\begin{equation}
	\label{eq:generalized_surrogate_fbeta_loss}
	\calL_{F_\beta,q} (\bff(\bx; \btheta), y)
	= y \cdot \frac{1 - f (\bx; \btheta)^q}{q} + 
	(1-y) \cdot \frac{ \left( \beta^2 \cdot \frac{p}{1-p} + f(\bx; \btheta) \right)^q - 1}{q}
	,
	\end{equation}
where the parameter $0< q \leq 1$ controls the level of noise robustness between MAE loss and BCE loss. 
That is, we can show that the limit 
$\lim_{q \rightarrow 0} \calL_{F_\beta,q} (\bff(\bx; \btheta), y)
= \calL_{F_\beta} (\bff(\bx; \btheta), y)$ approaches the surrogate $F_\beta$ loss, and
setting $q=1,\beta=1,p=0.5$ leads to the MAE loss.

%\subsection*{Acknowledgements}
%This work was supported by a National Research Foundation of Korea (NRF) grant funded by the Korea government (MSIT) (No. 2017R1C1B5076912).

\bibliographystyle{plainnat}
\bibliography{surrogate_loss}

\end{document}